\documentclass[journal, 11pt, onecolumn, draftclsnofoot]{IEEEtran}
\usepackage{epsfig}
\usepackage{amssymb}
\usepackage{float}  
\usepackage{array}
\usepackage[nodots]{numcompress}
\usepackage{graphicx} 
\usepackage{subfigure}
\usepackage{epsfig} 
\usepackage{amsmath} 
\usepackage{amssymb}  
\usepackage{cite}
\usepackage{url}
\usepackage{float}
\usepackage{graphicx}
\usepackage[misc]{ifsym}
\usepackage[colorlinks]{hyperref}
\hypersetup{colorlinks=true,citecolor=blue,linkcolor=blue,urlcolor=blue}
\usepackage[justification=centering]{caption}

\usepackage[ruled,linesnumbered]{algorithm2e}
\usepackage{booktabs}
\usepackage{multirow}
\usepackage{amsmath}
\usepackage{tabularx}
\usepackage{eurosym}
\usepackage{algpseudocode}
\algnewcommand\algorithmicswitch{\textbf{switch}}
\algnewcommand\algorithmiccase{\textbf{case}}
\algnewcommand\algorithmicassert{\texttt{assert}}
\algnewcommand\Assert[1]{\State \algorithmicassert(#1)}%
\algdef{SE}[SWITCH]{Switch}{EndSwitch}[1]{\algorithmicswitch\ #1\ \algorithmicdo}{\algorithmicend\ \algorithmicswitch}%
\algdef{SE}[CASE]{Case}{EndCase}[1]{\algorithmiccase\ #1}{\algorithmicend\ \algorithmiccase}%
\algtext*{EndSwitch}%
\algtext*{EndCase}%

\usepackage{multirow}
\usepackage{diagbox}
\newcommand{\tabincell}[2]{\begin{tabular}{@{}#1@{}}#2\end{tabular}}
\newcolumntype{N}{@{}m{0pt}@{}}

\title{\LARGE \bf
Lidar for Autonomous Driving: The principles, challenges, and trends for automotive lidar and perception systems
}
\author{You Li, Javier Ibanez-Guzman 
 \thanks{You Li, Javier Ibanez-Guzman are with research division at RENAULT S.A.S, 1 Avenue du Golf, 78280 Guyancourt, France
         {\tt\small you.li@renault.com, javier.ibanez-guzman@renault.com}}
}

\begin{document}
\maketitle
\thispagestyle{empty}
\pagestyle{empty}

\begin{abstract}
Autonomous vehicles rely on their perception systems to acquire information about their immediate surroundings. It is necessary to detect the presence of other vehicles, pedestrians and other relevant entities. Safety concerns and the need for accurate estimations have led to the introduction of Light Detection and Ranging (LiDAR) systems in complement to camera or radar based perception systems. This article presents a review of state-of-the-art automotive LiDAR technologies and the perception algorithms used with those technologies. LiDAR systems are introduced first by analyzing the main components, from laser transmitter to its beam scanning mechanism. Advantages/disadvantages and the current status of various solutions are introduced and compared. Then, the specific perception pipeline for LiDAR data processing, from an autonomous vehicle perspective is detailed. The model driven approaches and the emerging deep learning solutions are reviewed. Finally, we provide an overview of the limitations, challenges and trends for automotive LiDARs and perception systems.
\end{abstract}

\section{Introduction}\label{sec::introduction}
\begin{figure}[t]
\centering
\includegraphics[width = 0.5\textwidth, height=4cm]{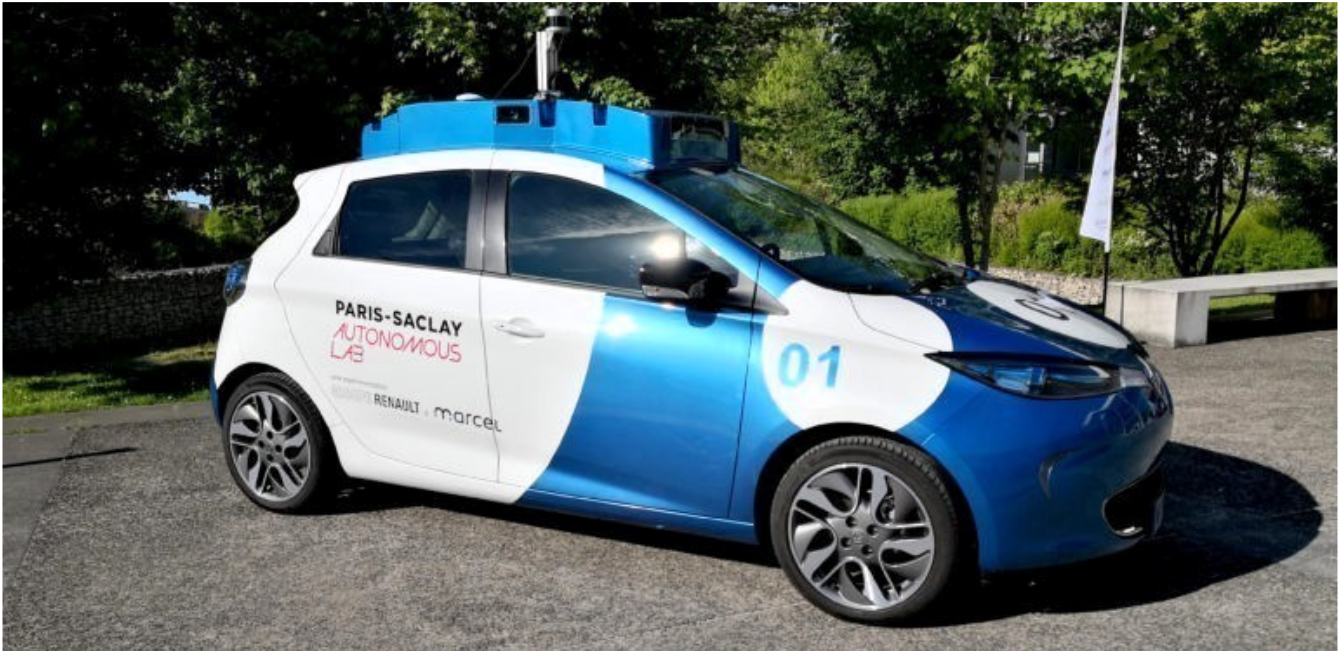}
\caption{The autonomous vehicle prototype developed by Groupe RENAULT for EVAPS project. The most evident sensor is the Velodyne UltraPuck LiDAR on top.}
\label{fig::pegasus2}
\end{figure}

Autonomous driving is entering a pre-industrialization phase with significant progress attained over the past years. Sensors initially capture data representation of the environment that are processed by perception algorithms to build the vehicle's immediate environment used for autonomous vehicle navigation. A perception system for autonomous vehicle navigation would consist of a combination of active and passive sensors, namely, cameras, radars, and LiDARs (Light detection and ranging) \cite{PerceptionJFR2008}. LiDARs are active sensors that illuminate the surroundings by emitting lasers. Ranges are measured precisely by processing the received laser returns from the reflecting surfaces. As stated in \cite{NRC2014}, LiDARs \textit{"poised to significantly alter the balance in commercial, military, and intelligence operations, as radar has done over the past seven decades"}. Despite much progresses in camera-based perception, image processing methods estimate distances. This approach posses difficulties when estimating distances for cross-traffic entities, in particular for monocular solutions. The 2007 DARPA Grand Challenge, a milestone in autonomous driving demonstrated the potential of LiDAR perception systems. The top 3 teams were all equipped with multiple LiDARs. A 64 layers LiDAR -- Velodyne HDL64 \footnote{\url{https://www.velodynelidar.com/hdl-64e.html}}, played as a critical central role for the champion team and runner-up \cite{Chris2008, Anna2009}. Currently, most high level autonomous vehicles use LiDARs as part of their perception systems despite their high cost and moving parts. A typical example is their incorporation into the autonomous vehicles being tested as robo-vehicles in different countries, e.g. the EVAPS field operational test in France\footnote{\url{https://pole-moveo.org/projets/evaps/}}, several participants (e.g. Renault, Transdev, Vedecom, etc) collaborate together to exploit mobility services based on autonomous vehicles in Saclay area in Paris. One of the prototypes developed by Renault is shown in Fig. \ref{fig::pegasus2}. As a result, there are more than 20 companies developing distinctive LiDAR systems for autonomous driving systems ranging from low level to high end, a sort of "big bang". Which LiDAR type(s) will dominate autonomous driving in the future is still in the mist.

On the other hand, LiDAR based algorithms also entered into a fast-track. For an autonomous vehicle, LiDARs are mainly used for \textit{perception} and \textit{localization}. Due to the page limit, this paper focuses only on perception usages. In the context of autonomous driving, a perception system provides a machine interpretable representation of the environment around the vehicle. From a user's perspective, the output of a perception system comprises 3 levels of information: 
\begin{itemize}
\item \textit{Physical description}: pose, velocity, shape of objects.
\item \textit{Semantic description}: the categories of objects.
\item \textit{Intention prediction}: likelihood of an object's behavior. 
\end{itemize}
Therefore, the LiDAR outputs are used for \textit{object detection, classification, tracking} and \textit{intention prediction}, corresponding to the various layers of information. Due to LiDAR's superiority in ranging accuracy, the provided physical information is highly reliable. While the semantic information carried by LiDAR is less, or more difficult acquired than camera -- a contextual sensor good at object recognition. In practice, LiDARs are combined with cameras to complement each other \cite{youlithesis}: camera is poor in distance estimation while LiDAR is not-satisfying for object recognition. Precise physical and semantic information, together with map information will improve the intention prediction without doubts. With many years of progresses, LiDAR-centric perception system become mature for model based processing algorithms. While emerging deep learning (DL) methods are changing this domain. Traditional model based LiDAR data processing methods are computation-friendly, explicable. While data-driven DL methods have demonstrated extraordinary capabilities in providing semantic information, which is the weak point of traditional methods.     

In this article, section \ref{sec::lidar_technologies} firstly introduce different ranging principles for a LiDAR, and dismantles its principal components. Several representative LiDAR products or manufactures are listed and classified according to the  different technical solutions. Section \ref{sec::lidar_perception} surveys the LiDAR data processing algorithms, including classic methods and the emerging deep learning approaches, to provide different level of information required from a perception system. Section \ref{sec::conclusion} covers open challenges for the current LiDARs and LiDAR based perception systems for the automotive industry, and their future trends according to the authors' opinions.

\section{LiDAR technologies}\label{sec::lidar_technologies}
\begin{figure}[t]
\centering
\includegraphics[width = 0.8\textwidth]{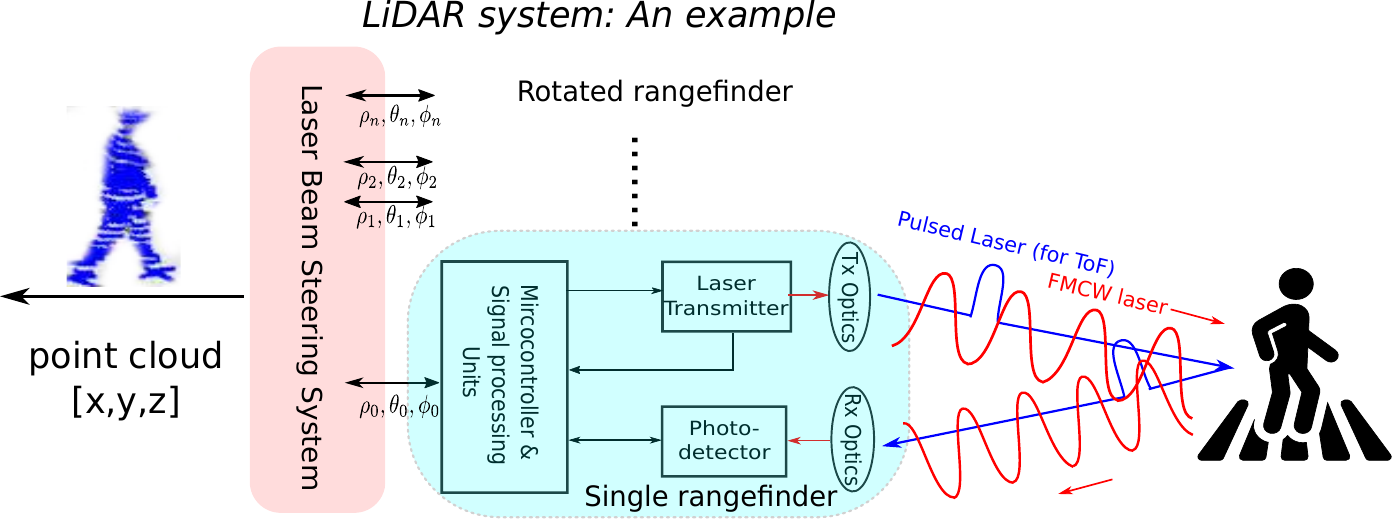}
\caption{An example of ToF laser rangefinder. The rangefinder uses either direct or coherent method to measure the distance at a certain direction controlled by the scanning system.}
\label{fig::lidar_system}
\end{figure}

A typical LiDAR operates by scanning its field of view with one or several laser beams. This is done through delicately designed beam steering system. The laser beam is generated by an amplitude modulated laser diode that emits at near infrared wavelength. The laser beam is reflected by the environment back to the scanner. The returned signal is received by photodetector. Fast electronics filter the signal and measure the difference between the transmitted and received signals that proportional to the distance. Range is estimated from the sensor model based on this difference. Difference in variations of reflected energy due to surface materials as well as state of the milieu between transmitter and receiver are compensated through signal processing. The LiDAR outputs comprise both 3D point clouds that corresponds to the scanned environments and the intensities that correspond to the reflected laser energies. Figure \ref{fig::lidar_system} shows a conceptual representation of this operating principle.

A LiDAR system can be partitioned into: the \textit{laser rangefinder} system and the \textit{scanning system}. The laser rangefinder comprises the \textit{laser transmitter} that illuminates the target via a modulated wave; the \textit{photodetector} that generates electronic signal from the reflected photons after optical processing and photoelectric conversion; \textit{optics} which collimate the emitted laser and focus the reflected signal onto the photodetector, and \textit{signal processing electronics} to estimate the distance between the laser source and the reflecting surface, based on the received signal. The \textit{scanning system} typically will steer laser beams at different azimuths and vertical angles, denoted by $\phi_i, \theta_i$, where \textit{i} is an index that determines the direction at which the beam is being pointed.

This section addresses initially the principles of a rangefinder in order to understand the measurement process and limitations, then it introduces the scanning systems that define the sensor field of view. It is then possible to classify LiDARs based on the technologies they use. This classification is then applied to examine the commercially available automotive type LiDARs.

\subsection{Laser Rangefinder Principles}
A rangefinder that measures the distance to an object by a laser beam is known as a laser rangefinder. The manner in which they operate depends on the type of signal modulation used in the laser beam. Pulsed Laser are used so their\textit{time-of-flight (ToF)} can be measured, these are known as \textit{direct detection} laser rangefinders. The laser signal can be also a \textit{frequency-modulated continuous wave (FMCV)} that indirectly measures both distance and velocity, from the Doppler effect. These are known as \textit{coherent detection} laser rangefinders. 

\subsubsection{LiDAR power equation}
The transmitted laser is firstly attenuated through the transmission medium, then diffused as it reflects from the target surface. It is partially captured by the receiving optics and finally transformed into an electrical signal by a photodetector. For a target at distance $r$, the amount of received power $P_r$ by the photodetector based from a pulsed laser emitter can be approximately modeled as\cite{IntroLiDAR2005}:
\begin{equation}
  P_r = E_p\frac{c\eta A_r}{2r^2} \cdot \beta \cdot T_r
  \label{eq::lidar}
\end{equation}
where $E_p$ is the total energy of a transmitted pulse laser, $c$ is light speed. $A_r$ represents the area of receive aperture at range $r$. $\eta$ is the overall system efficiency. $\beta$ is the reflectance of the target's surface, which is decided by both surface properties and incident angle. In a simple case of Lambertian reflection with a reflectivity of $0<\Gamma<1$, it is given by: $\beta = \Gamma/\pi$. The final part $T_r$ denotes the transmission loss through the transmission medium. When the LiDAR works under adverse conditions (e.g. fog, rain, dust, snow, etc), the particles within the air would scatter and absorb the photons. Eq. \ref{eq::lidar} reveals that the received power $p_r$ decreases quadratically with respect to distance $r$: an object at hundreds meters away is order of magnitudes "darker" than at tens of meters. Simply increasing the power of laser transmitter is restricted by the eye-safety standard: IEC 60825\cite{IEC60825}. To overcome this, the overall system efficiency need to be improved through optics, photodetectors, and more advanced signal processing algorithms. For a FMCW laser, Eq. \ref{eq::lidar} still holds except slight differences.


\subsubsection{Time-of-flight (ToF)}
A ToF laser rangefinder measures the range by calculating the time difference between the transmitted and received lasers:
\begin{equation}
  r = \frac{1}{2n}c\Delta t
  \label{eq::tof}
\end{equation}
where $c$ is the light speed, $n$ is the index of refraction of the propagation medium ($\approx$ 1 for air). $\Delta t$ is the time gap between the transmitted laser and the received laser. ToF LiDARs prevail in current automotive LiDARs market as their simple structure and signal processing methods. However, potential for increasing their maximum range is constrained by limited transmit power due to eye-safety requirements. In ToF LiDAR, the return signal can be interference from strong sunlight or laser beams from other ToF LiDARS.


\subsubsection{Coherent detection}

\begin{figure}[t]
\centering
\includegraphics[width = 0.4\textwidth]{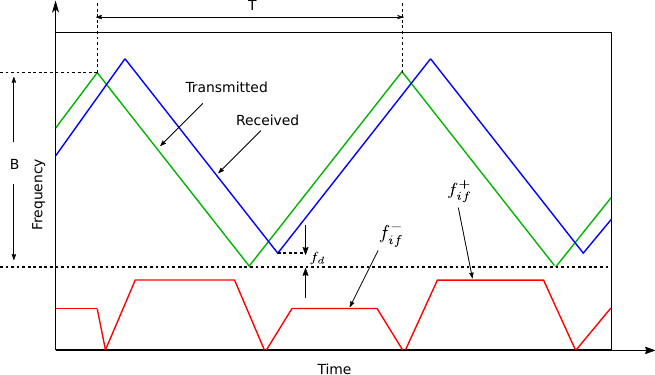}
\caption{Principle of coherent detection: distance is estimated by the intermediate frequency (in red trace) generated by mixing transmitted and received light waves (from \cite{nasa2008}).}
\label{fig::range_principle}
\end{figure}

By mixing the local carrier signal with the received signal, it is possible to demodulate the received signal and thus it is possible to obtain the phase and frequency shift of the laser signal, hence acquire the distance and velocity from the reflecting surface. This can be regarded as the optical version of FMCW (Frequency Modulated Continuous Wave) Radars, which are popular nowadays passenger vehicles' ADAS systems. 
FMCW LiDARs continuously emit a FM laser signal (e.g. linearly-chirped laser) to a target while keeping a reference signal, also known as a local oscillator. The common modulation functions are sawtooth or triangle waves. Because a FMCW LiDAR continuously illuminates the objects using less emitted power for this purpose, complying with eye-safe requirements and opening the possibility to use more power to extend their field of view. 
The signals used for coherent detection are shown in Fig. \ref{fig::range_principle}. The intermediate frequency (IF, red trace) can be generated by mixing the local oscillator signal (a linearly-chirped triangular modulation function) from the laser transmitter (in green) with the laser signal reflected from observed surfaces (in blue). Processing of the resulting signal generates an intermediate frequency $f_{if}$ shown in red in Fig. \ref{fig::range_principle} (b). By assuming that the Doppler frequency shift $f_d$ is less than the intermediate frequency $f_{if}$, we have:
\begin{equation}
f_{if} = \frac{4rB}{ct}= \frac{f_{if}^+ + f_{if}^-}{2}, \quad f_{d} = \frac{f_{if}^+ - f_{if}^-}{2} (when \quad f_d<f_{if})
\end{equation}
where $B, r, t$ are the modulation bandwidth, the waveform period and the light speed, respectively. The velocity is obtained as:
\begin{equation}
v = \frac{f_d \lambda}{2}
\end{equation}
where $\lambda$ is the laser wavelength

FMCW LiDAR is able to directly measure the distance and velocity at the same time, while for ToF LiDAR, speed is obtained from indirect estimation through several consecutive sensor readings. By using a FMCW laser signal, it is possible to reduce the interference effect from other laser sources and strong sunlight. However, FMCW LiDARs require high quality laser generators that possess long coherent distance.  

\subsection{Laser Transmission and Reception}
The generation of the laser signals and their emission as well as the receiver electronics of the reflected signals also characterise the performance and cost of the laser rangefinders.
\subsubsection{Laser sources}
ToF LiDARs need a pulsed (amplitude modulated) laser signal. This is generated using a pulsed laser diode or a fiber laser. A diode laser causes laser oscillation by flowing an electric current to the diode's junction. Diode lasers can be grouped into 2 classes: \textit{Edge-emitting laser (EEL)} and \textit{Surface-emitting semiconductor laser (VCSEL)}. EEL have been applied in the telecommunication industry for a long time. VCSEL outputs a circular beam, while EEL transmits an elliptical laser beam, requiring additional beam-shaping optics. In VCSEL, forming a 2D laser array on a single chip is easier than for an EEL, this is important as it increases the LiDAR resolution. By contrast, the range for VCSEL is shorter due to power limits. The pulsed laser diodes using in automotive applications are hybrid devices. That is, a laser chip is mounted with capacitors that are triggered by a MOSFET transistor. Thus at every gate opening, the electric charge accumulated in the capacitors will be discharged into the chip, this will emit the optical pulse in a controlled manner. These sources are cost-effective, as their 905 nm output can be detected by cost effective silicon detectors. However, these diodes have a limited pulse repetition rate and lower peak power and might require cooling. Laser diode sources for 3D flash LiDAR use diode stack technology with several edge-emitting bars assembled into a vertical stack. Heat dissipation becomes an issue, hence the need for heat sinks as well as the accumulation of emitted power beyond eye-safe requirements. Fiber laser can have higher output power which is useful when operating at high wavelengths. Their output beams can be split and routed to multiple sensor locations using optical fiber, they have better pulse repetition frequency, better beam quality, etc. However, they can be bulky and thus resulting in non-compact systems that are difficult to be integrated in vehicles.

\subsubsection{Laser wavelength}
Selecting an appropriate wavelength of laser should have a comprehensive consideration of atmospheric windows, eye-safety requirement and the cost. The 850-950nm (near infrared (NIR)) and 1550nm (short wave infrared (SWIR)) lasers are mostly utilized because of their popularities in industry. Either a low price diode laser or a more powerful fiber laser at wavelength 850-950nm or 1550nm is easily purchased from the market. The maximum power permitted by eye-safety standard for 1550nm laser is higher than the lasers in 850-950nm, which means a larger range could be achieved. However, expensive InGaAs based photodiode is required to detector laser returns at 1550nm. The efficiency of InGaAs based photodiode is lower than the matured Silicon ones for NIR lasers. In addition, the atmospheric water absorption for 1550nm is stronger than 850-950nm. Therefore, LiDAR systems at NIR wavelength (905nm for instance) are still the mainstream.   

\subsubsection{Photo-detector}
A photo-detector converts optical power to electrical power by the photoelectric effect. Photo-sensitivity that describes a photodetector's response when receiving photons is one of the most critical characteristics. The photosensitivity depends on the wavelength of the received laser. Therefore, selecting photodetector for a LiDAR system is closely related to the choice of laser wavelength. The most popular detectors are \textit{PIN photodiode, Avalanche photodiode (APD), Single-photon avalanche diode (SPAD), Silicon Photomultiplier (SiPM).} 

 \textit{PIN photodiode:} is formed by a p-i-n junction that creates a depletion region that is free of mobile charge carriers. By applying a reverse bias to a photodiode, absorbing a photon will generate a current flow in the reverse-biased photodiode. 

 \textit{Avalanche Photodiode (APD):} is a photodiode that applies reverse voltage to multiply photocurrent through avalanche effect. The APD's ability to multiply signals reduces the effect of noise and achieves higher internal current gain (around 100) and SNR than the PIN photodiode. Therefore, APDs are quite common in contemporary LiDAR systems. Silicon based APDs are sensitive through the visible spectral region until the near infrared around 1000nm. At longer wavelengths up to 1700nm, InGaAs APDs are available, although at higher cost. 

 \textit{Single-photon avalanche diode (SPAD):} is an APD designed to operate with a reverse-bias voltage above the breakdown voltage (Geiger-mode), which allows a detection of very few photons in very short time. SPAD can achieve a gain of $10^6$ that is significantly higher than APD. This characteristic allows the SPAD detect extremely weak light at long distance. Furthermore, the CMOS technology that can be used for the SPAD fabrication enable an integrated array of photodiodes on one chip. This is desirable for increasing LiDAR's resolution while cut the cost and power consumption.  

 \textit{Silicon Photomultiplier (SiPM):} is based on SPAD, while enable photon counting. The Geiger-mode in which a SPAD operates is a photon-trigger mode that a SPAD cannot distinguish the magnitude of received photo flux. To overcome this issue, SiPM integrates a dense array of 'microcells' (a pair of a SPAD and a quench resistor) working identically and independently. SiPM's output is in essence a combination of the photocurrents detected from each microcell. In this approach, SiPM is capable of giving information on the magnitude of an instantaneous photon flux.

\subsection{Scanning system}
\label{sec::scanning}
A scanning system (or beam steering system) is designed to enable the transmitted lasers to rapidly explore a large area. The existing scanning approaches are usually classified either \textit{mechanical spinning} or \textit{solid state}. The former usually contains a bulky rotating mirror system like the Velodyne HDL64 in the early ages of autonomous driving history. The later \textit{"solid state"} refers to scanning system without moving parts (even some are still steered by micro-mirrors), what is preferred by the automotive industry.


\begin{figure*}[t]
  \centering
  \subfigure[Principle of mechanical spinning LiDAR]{
\includegraphics[width = 0.4\textwidth]{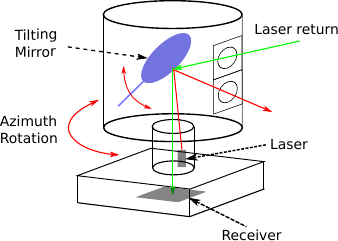}
}
  \subfigure[Principle of MEMS LiDAR]{
    \includegraphics[width = 0.4\textwidth]{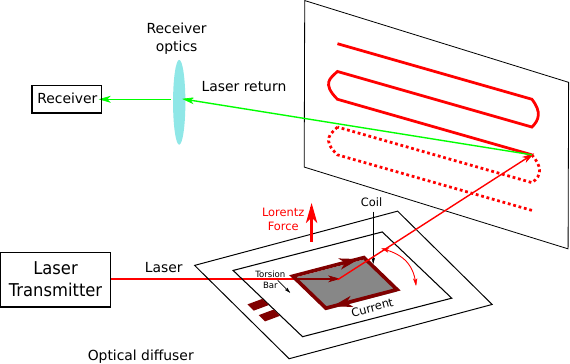}
  }
  \subfigure[Principle of flash LiDAR]{
    \includegraphics[width = 0.4\textwidth]{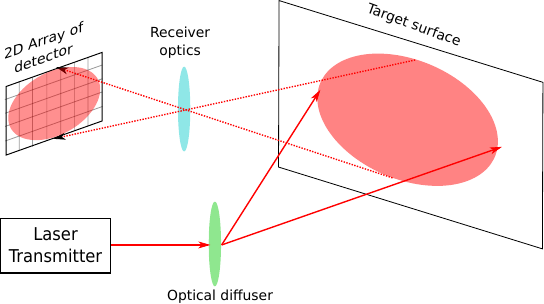}
  }
    \subfigure[Principle of OPA LiDAR (from \cite{hamatasu2017})]{
    \includegraphics[width = 0.4\textwidth]{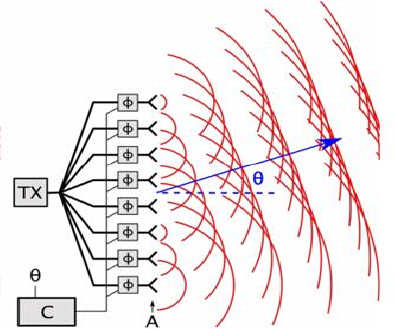}
  }
\caption{LiDAR systems categorized by scanning approaches.}
\label{fig::scanning_system}
\end{figure*}

\begin{figure*}[t]
  \centering
  \subfigure[]{
\includegraphics[width = 0.25\textwidth, height = 4cm]{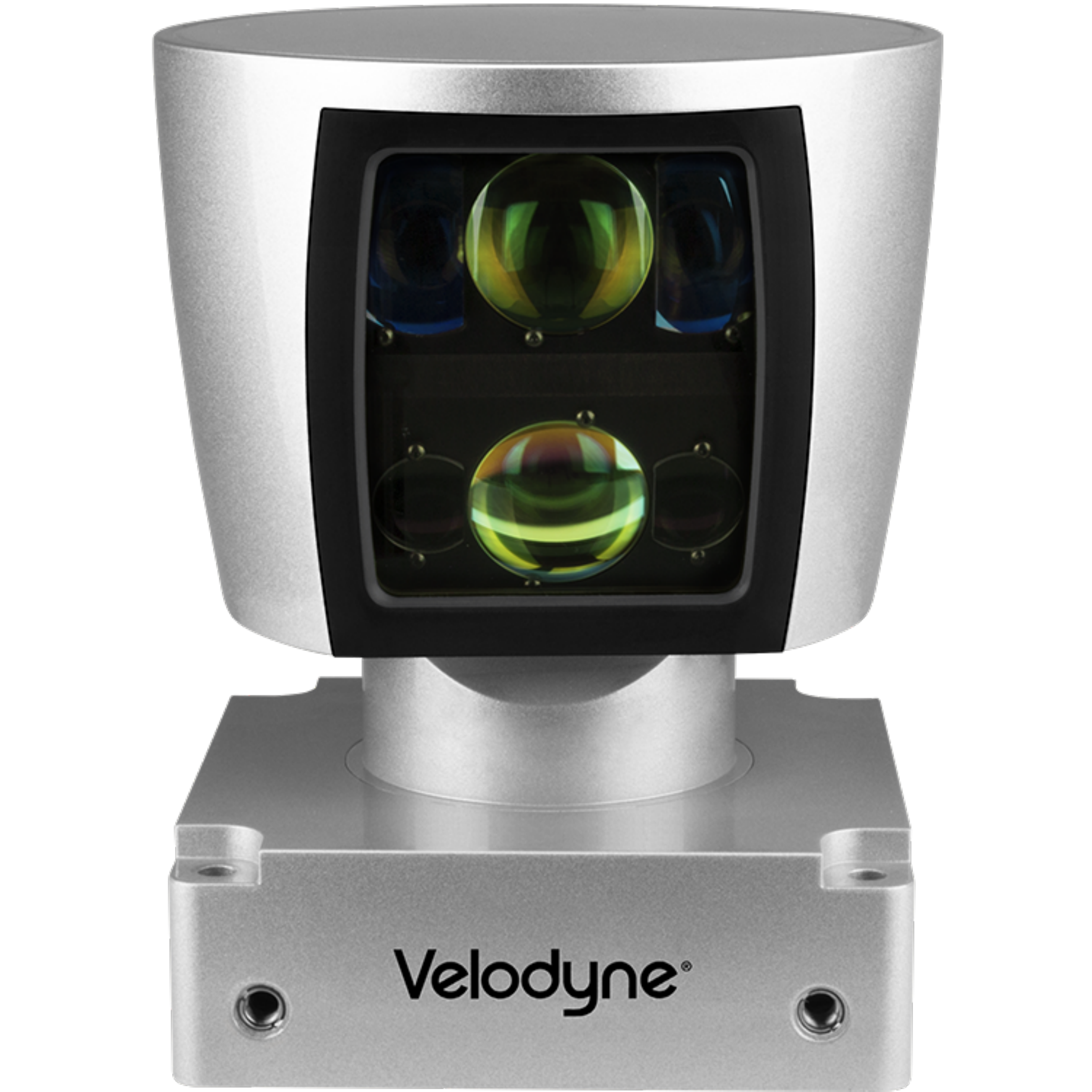}
}
  \subfigure[]{
    \includegraphics[width = 0.28\textwidth,height = 4cm]{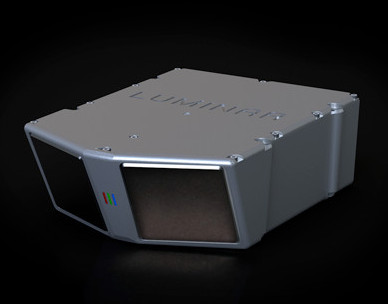}
  }
  \subfigure[]{
    \includegraphics[width = 0.3\textwidth, height = 3cm]{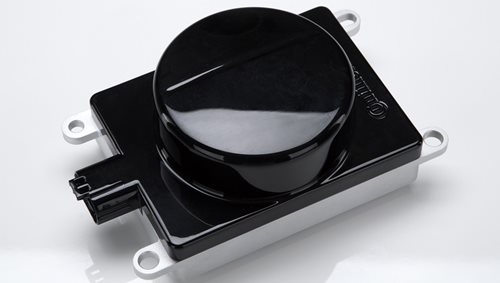}
  }
\caption{LiDAR product examples: (a) mechanical spinning 905nm LiDAR from Velodyne, (b) 1550nm MEMS LiDAR from Luminar (c) Flash LiDAR from Continental}
\label{fig::lidar_examples}
\end{figure*}


\textit{Mechanical Spinning:} The currently most popular scanning solution for automotive LiDAR is the mechanical spinning system\cite{depth2016}, which steers the laser beams through rotating assembly (e.g. mirror, prism, etc) controlled by a motor to create a large field of view (FoV). Conventionally, nodding-mirror system and polygonal-mirror system \cite{Oneill2011} are the main types applied. For example, as the mechanical spinning scheme shown in Fig. \ref{fig::scanning_system} (a), An embedded nodding-mirror system tilts the lasers to generate a vertical FoV. Then, $360^\circ$ horizontal FoV is achieved by rotating the LiDAR base. State-of-the-art LiDAR use multiple beams to reduce the movable mechanism. For instance, Velodyne VLP series use arrays of laser diodes and photodiodes to increase point cloud densities. The mechanical spinning system offers an advantage of high signal-to-noise ratio (SNR) over a wide FOV. However, the rotating mechanism is bulky for integration inside a vehicle, and is fragile in harsh conditions such as vibration, what is quite common in automotive applications. A typical product example: Velodyne's HDL64 is shown in Fig. \ref{fig::lidar_examples} (a). 

\textit{MEMS Micro-Scanning:} MEMS (Micro-Electro-Mechanical Systems) technology allows the fabrication of miniature mechanical and electro-mechanical devices using silicon fabrication techniques. In essence, a MEMS mirror is a mirror embedded on a chip \cite{MEMS2018}. The MEMS mirror is rotated by balancing between two opposite forces: an electromagnetic force (Lorentz force) produced by the conductive coil around the mirror, and an elastic force from torsion bar, which serves as the axis of rotation. This principle is shown in Fig. \ref{fig::scanning_system} (b). The MEMS mirrors can be either single-axis for 1D movement \cite{mems1D2018}, or dual-axis for 2D movement. Also, MEMS mirror can work in resonant mode at its characteristic oscillation frequency to obtain a large deflection angle and high operating frequency. In non-resonant mode, MEMS scanning mirror can be controlled to follow programmed scan trajectory. For example, for MEMS based AEYE LiDAR, the LiDAR can dynamically change the FoV and scanning path in order to focus on some critical parts. Although MEMS LiDARs still contain moving parts, this near-solid-state technology is still promising because of the mature techniques in IC industry are able to satisfy the strict cost requirements. An example of Luminar's MEMS 1550nm LiDAR is shown in Fig. \ref{fig::lidar_examples} (b).

\textit{Flash:} Originally applied for spacecraft in autonomous landing and docking with satellites, 3D flash LiDARs \cite{AIAA2016} totally remove the rotating parts within scanning systems. Hence, they are truly solid-state. A flash LiDAR behaves as a camera. A single laser that is spread by an optical diffuser to illuminate the whole scene at once. Then, it uses a 2D array of photodiodes (similar to the CMOS/CCD for camera) to capture the laser returns, which are finally processed to form a 3D point clouds, as shown in Fig. \ref{fig::scanning_system} (c). Since all the pixels of flash LiDAR measure the ranges simultaneously, the issue of movement compensation caused by platform motion is avoided. In addition, the semiconductor-based 3D flash LiDARs facilitate fabrication and packaging for massive production that lead to lower cost. However, the critical issue of 3D flash LiDAR is its limited detecting range (usually $<100m$) because a single diffused laser is responsible for detecting the whole area under a small power threshold for eye-safety. Another disadvantage is the limited FoV because it cannot rotate and scan the surroundings like a scanning-type LiDAR does. A typical example of Continental's commercial flash LiDAR product for middle range perception is shown in Fig. \ref{fig::lidar_examples} 

\textit{OPA (Optical Phased Array):} As a type of true solid-state LiDAR, optical phased array (OPA) LiDARs \cite{Poulton2018} \cite{paul1996} don't comprise moving components. Similar to the phased array Radar, an OPA is able to steer the laser beams through various types of phase modulators. The speed of light can be changed by the optical phase modulators when the lasers are passing through the lens, as illustrated in Fig. \ref{fig::scanning_system} (d). Consequently, different light speeds in different paths allow control of the optical wave-front shape and hence the steering angles. Although OPA had been placed high hopes as a promising technology, there is not a commercial product yet in the market.  


\begin{table}[t]
  \small
  \centering
  \begin{tabular}{c|c|c|c|c|c|c}
    \toprule
    \multicolumn{2}{c|}{} &Mechanical Spinning &MEMS &Flash &OPA & Undisclosed\\\hline\hline
    \multirow{2}{*}{\tabincell{c}{ToF LiDAR}}&NIR &\tabincell{c}{Velodyne, IBEO, \\Valeo, Ouster*\\Hesai, Robosense}&\tabincell{c}{Innoviz\\Robosense} &\tabincell{c}{Continental\\ Xenomatix} &Quanergy & \\\cline{2-7}
    &SWIR &Luminar &\tabincell{c}{AEYE\\Hesai} & \tabincell{c}{Argo*\\(Princeton Lightwave)} & & \\\hline
    \multicolumn{2}{c|}{\tabincell{c}{FMCW LiDAR}}& & & &\tabincell{c}{Cruise\\(Strobe)}  &\tabincell{c}{Aurora\\(Blackmore,1550nm)}\\
    \bottomrule
  \end{tabular}
  \caption{Representative LiDAR manufactures and the adopted technologies. The manufactures marked by * utilize single photon Geiger-mode SPAD as photodetector.}
  \label{tab::lidar_players}
\end{table}

\subsection{Current Status of Automotive LiDAR }
Mechanical spinning LiDAR is the first stepping into mass-produced car. Announced in 2017, Audi released its latest luxury sedan A8 that equipped a Valeo's Scala LiDAR for automated driving functions -- the first commercial available vehicle carrying automotive grade LiDAR in the world. Valeo's scala \footnote{\url{https://www.valeo.com/en/valeo-scala/}} is a 4 layer mechanical spinning LiDAR similar to its cousin, IBEO Lux4. Empowered by Scala, A8 is able to achieve L3 level automated driving functions, without the need of hands on the steering wheel (need to be allowed by legislation). In 2019, VALEO gained a \euro500M order of its next generation LiDAR -- Scala2, from several car manufacturers.

In the same time, to reduce the cost and to improve the robustness, plenty of companies focus on solid state scanning system. As shown in Tab. \ref{tab::lidar_players}, Innoviz, Continental and Quanergy are developing MEMS, Flash and OPA LiDARs, respectively. In 2018, BMW announced the collaboration with Innoviz for series production in 2021. To increase the maximum detection range, some used SPAD array working in single-photon detection mode (Geiger-mode). Ouster OS-1 64 \footnote{\url{https://www.ouster.io/blog-posts/2018/11/8/how-multi-beam-flash-lidar-works}} adopted CMOS based SPAD to detect 850nm lasers emitted by 2D VCSEL laser array. Toyota made a LiDAR prototype containing a CMOS SPAD array (202x96 pixels) for receiving 905nm lasers\cite{toyotaSPAD2016}. Princetion Lightware (acquired by Argo.ai) also realized a SPAD LiDAR prototype \footnote{\url{https://www.youtube.com/watch?v=K2YkrFSJc3A}}, while few information is disclosed. As for SiPM, a LiDAR prototype \cite{sipm2017} has been made by SensL (acquired by OmmniVision), while the commercial products are still under development.

Some switch to SWIR laser (e.g. 1550nm) because of the higher power threshold allowed than NIR laser, such as Luminar (announced collaboration with Toyota), AEYE. Coherent detection based FMCW LiDARs are chased by car manufactures and investments as well. Strobe and Blackmore, the two representative FMCW LiDAR start-ups, are quickly acquired by Cruise and Aurora, respectively. In \ref{tab::lidar_players}, we classify and list several representative automotive LiDAR suppliers and their disclosed technologies.

Another trend of LiDAR is to overcome the adverse weather conditions, such as rain, fog, snow, dust, etc \cite{lidarsnow2015,matti2018}. According to Eq. \ref{eq::lidar}, the adverse weather conditions increase the transmission loss $T_r$ and weaken the reflectivity of an object $\beta$ that the received energy becomes less. Because SWIR laser (e.g. 1550nm) can achieve higher transmission power, LiDARs belonging to this wavelength are expected to have better performances in harsh weather.

\section{LiDAR Perception System}\label{sec::lidar_perception}
\begin{figure}[t]
  \centering
\includegraphics[width = 0.9\textwidth]{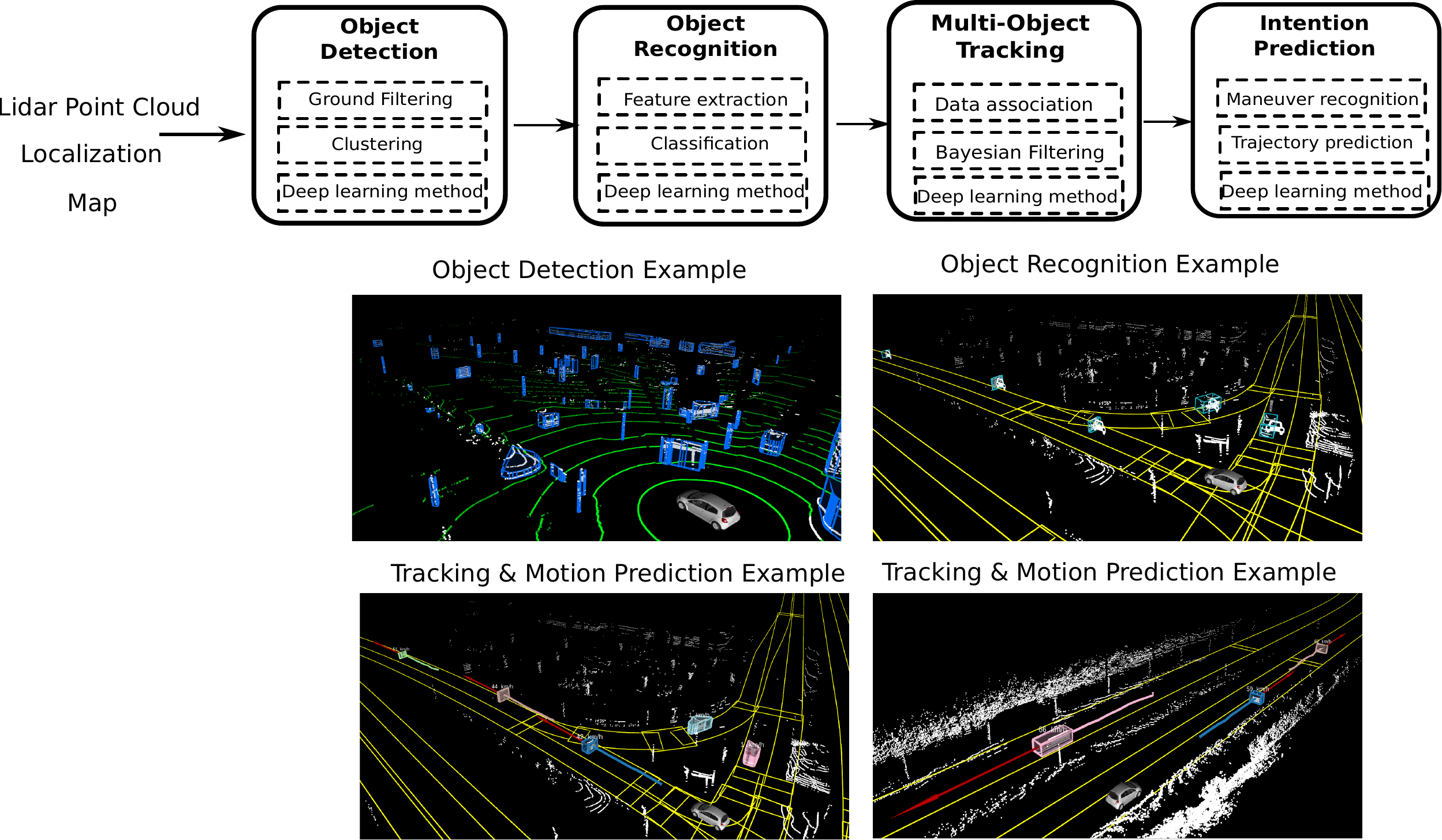}
\caption{Pipeline of a classic LiDAR perception system, with exemplary outputs of each step. The examples are from authors' platform as shown in Fig. \ref{fig::pegasus2}. Please note that after object detection, we only process the objects within the road (as denoted by the yellow lines).}
\label{fig::perception}
\end{figure}

For an autonomous vehicle, its perception system interprets the perceived environment into hierarchical object descriptions (i.e. physical, semantic, intention awareness) from the perception sensor outputs, localization and map data. As shown in Fig. \ref{fig::perception}, traditional pipeline \cite{himmel2008,Anna2009} of processing LiDAR data consists of 4 steps: \textit{object detection}, \textit{tracking, recognition} and \textit{motion prediction}. The recent rise of deep learning technologies is changing this classic flow that we will introduce it after the classic approaches. Due to the popularity of Velodyne LiDARs in research communities, the reviewed data processing methods are mainly based on this mechanical spinning LiDAR. 

\subsection{Object Detection}
The object detection algorithms extract the object candidates and estimate their physical information: the positions and shapes of the detected objects. Since in most traffic scenes, targets are perpendicular to a flat ground, object detection algorithms usually comprise: \textit{Ground filtering} and \textit{Clustering}. \textit{Ground filtering} labels a point cloud either ground or non-ground. Then, non-ground points are grouped into different objects by \textit{clustering} methods.

In an early research\cite{Anna2009}, the point clouds from LiDAR are projected into polar grids subdividing $360^{\circ}$ around the LiDAR. The points inside each grid cell are treated consecutively to generate a virtual scan, which specifies the region of \textit{free, occupied} and \textit{occluded}. Occupied virtual scans are grouped into object clusters. \cite{himmel2008} followed this method while utilized grid-based local plane fitting approach instead of processing every point as in \cite{Anna2009}. The grids that are able to be fitted as a plane are classified as ground grids, the remainning non-ground grids are clustered by the connected component labeling (CCL). However, the polar grid based methods always need a projection of 3D LiDAR points into discrete grids, which loses raw information from LiDAR measurements. 

\begin{figure}[t]
  \centering
\includegraphics[width = 0.8\textwidth]{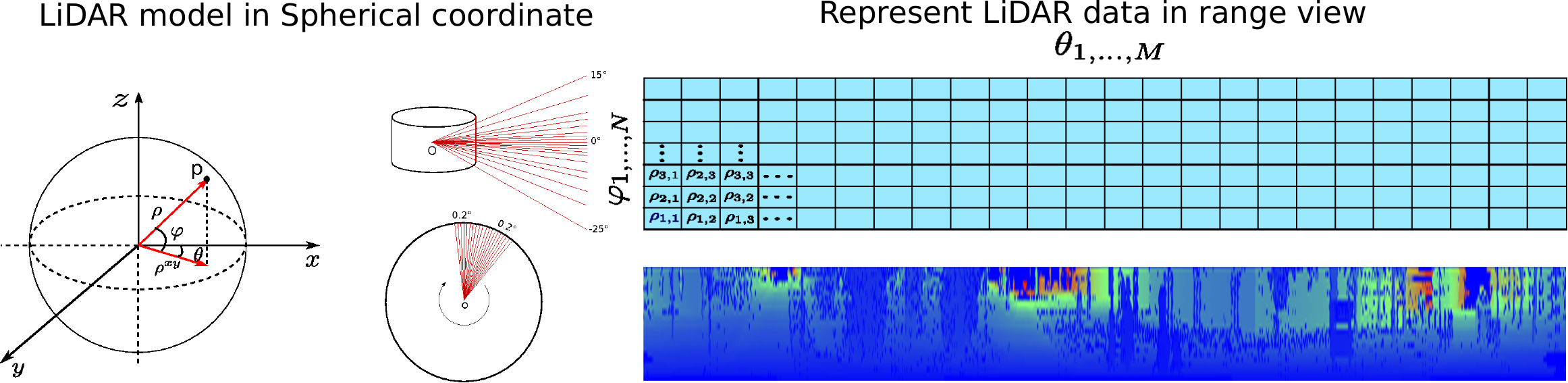}
\caption{Range view of spinning LiDAR (Velodyne UltraPuck) for further processing. The range image (32x1800) in pseudo-color facilitates the following processing.}
\label{fig::range}
\end{figure}

Processing LiDAR signals in spherical coordinates $(r,\varphi,\theta)$ provides a better approach. For the Velodyne UltraPuck used by the authors, vertical angle for each laser beam is fixed, azimuth angle is decided by the scanning time and motor speed. Therefore, every range reading can be represented by $P_{i,j} = (\rho_{i,j}, \varphi_i, \theta_j,)$, where $i$ refers to a certain laser beam, $j$ is the azimuth angle index, as seen in Fig. \ref{fig::range}. This approach naturally fills the range readings into a predefined data buffer (range image) and hence allows fast access to a point and its neighbours. Processing LiDAR data in range view is becoming popular in recent years. For instance, based on a range image, \cite{Igor2016} segmented the ground points in each column. The remaining non-ground points are grouped easily through criterions of distance and angle. For a 32 beams LiDAR, they reached 4ms in an Intel i5 processor. \cite{Dim2017} processed the range image row by row. They applied the clustering in each scan-line (actually the row in range image), and then merge the clusters scan-line by scan-line. Fig. \ref{fig::perception} shows a sample result of ground filtering and clustering from our implementation based on range image as well. The green points are ground points, non-ground points are grouped into object candidates (in blue polygons). Object detection provides an initial physical information, e.g. position of the an object. The following steps such as recognition and tracking complement semantic and more physical information e.g. heading, speed, to the detected objects.

\subsection{Object Recognition}
Machine learning based object recognition methods furnish the semantic information (e.g. classes of pedestrian, vehicle, truck, tree, building, etc) to the detected objects. A typical recognition process employed in \cite{himmel2008} comprises a \textit{feature extraction} step calculating compact object descriptors, and a \textit{classification} step where pre-trained classifiers predict the categories of objects based on the extracted features. As summarized in \cite{chen2014}, the features proposed in literature can be roughly divided into two classes: the \textit{global features} for the whole object, or the \textit{local features} for each point. An object's size, radius, central moments or the maximum intensity \cite{Cristiano2009, himmel2008} are the most basic global features. Applying Principal Component Analysis (PCA) in 3D point clouds is another effective method to acquire global shape features. As adopted in \cite{douillard2014}, three salience features (surfaceness, linearness, scatterness) can be acquired by analyzing the eigenvalues acquired from PCA. As for local feature, \cite{himmel2008} calculated the three salience features for each point and its neighboring points. 3 histograms, each containing 4 bins spaced between 0 and 1, for the three salience features are extracted as local features. A more complicated feature is the Spin Image (SI) introduced by \cite{spin1999}. A SI is created by spinning a grid around the surface normal $n$ of a given point $p$. The virtual pixels of a SI are the distances either to line through the $n$ or to the plane defined by $p$ and $n$. \cite{douillard2014} transformed this individual point-wise feature to global feature: for an object, only the SI of its central point is utilized as object descriptor. In literature, there are more sophisticated features, such as Global Fourier Histogram (GFH) \cite{chen2014} descriptor. However, real-time requirement restrains the complexity of features.

After feature extraction, the following classification is a typical supervised machine learning process: a classifier trained by a ground truth dataset predict the class of input objects. Well-known datasets such as KITTI \footnote{\url{http://www.cvlibs.net/datasets/kitti/}} provide abundant resources. From the arsenal of machine learning (ML), plenty of ML tools can be applied, such as Naive Bayes\cite{Cristiano2009}, Support Vector Machine (SVM)\cite{himmel2008, chen2014, zeng2012}, KNN \cite{douillard2014}, Random Forest (RF) and Gradient Boosting Tree (GBT) \cite{youliIV2012} can be applied. SVM with RBF (radial basis function) kernel is still the most popular method, considering its speed and accuracy. Fig. \ref{fig::perception} shows the recognition results on the detected on-road objects based on our implementations (SVM with RBF kernel). Recently, \cite{edouardIV19} applied an evidential neural network to classify the LiDAR object. Evidential classifier can better handle the \textit{unknown} classes that are frequently encountered in practice.

\subsection{Object Tracking}
Multiple object tracking (MOT) algorithms correlate and locate the detected/recognized objects through spatio-temporal consistency. MOT maintains the identities of detected objects and yields their physical states i.e. trajectories, poses, velocities. MOT is a classic engineering problem \cite{FOT2011} that has been researched for a long time. A basic architecture mainly comprises \textit{single object tracker} that "optimally" estimates the state of tracked object, \textit{data association} that assign the new detections to the trackers.

A \textit{single object tracker} models the movement as a dynamic state-space model and estimates the state under the Bayesian filtering framework. Kalman Filter (KF) family -- the classic KF under Gaussian-linear assumption and its variants Extended Kalman Filter (EKF), Unscented Kalman Filter (UKF), is the popular toolbox. \cite{kfLIDAR2013} employed a KF with a constant velocity model to track LiDAR detections. As a non-linear version of KF, EKF is utilized for LiDAR object tracking in \cite{ekfLiDAR2014}. Extending the single dynamic model to multiple maneuver models, the interacting multiple model (IMM) filter is able to handle more complicated cases. IMM filter consists of several filters running in parallel and each filter use different motion model. For a single object, an IMM-UKF filter was applied in \cite{tud2017}, where three UKFs work for three motion models: Constant Velocity, Constant Turn Rate and Random Motion.

As another common approach, Particle Filter (PF) is designed for more general cases that not meet the Gaussian-linear assumption. The application of PF in LiDAR data processing can be dated from DARPA Grand Challenge 2017 \cite{Anna2009}, where a Rao-Blackwellized Particle Filter is used. However, PF requires a large number of particles, especially for high dimensional state space. Hence, KF family is more popular in real-time perception systems.      

\textit{Data association} associates the detections with the tracks. The simplest method is the nearest neighbor (NN) filter (implemented in \cite{kfLIDAR2013}) which assigns the detections to their closest tracks, based on Euclidean or Mahalanobis distance between the detection and the track. NN filter is insufficient for clutter scenarios. In contrast, the joint probabilistic data association filter (JPDAF) offers a soft, probabilistic way for the detection-track association. JPDAF considers all the possible detections (including no detection) in a gating window and estimates their assignment probabilities to the tracks and takes the weighted average of all the association hypotheses. In \cite{tud2017}, JPDAF was applied for data association and an IMM-UKF tracker is to track an individual object. 

In contrast to the radar based MOT in which all the detections are usually modeled as points, LiDAR based MOT is distinctive in that it should track the shapes of detections as well. The simplest shape model is a 2D bounding box \cite{Anna2009}, which assumes the detections are car-like objects. L-shape fitting  \cite{Lfitting2017} is the most common approach estimating the bounding box's center, width, height and heading. However, 2D bounding box is insufficient for more general objects, such as pedestrian, tree, building etc. A more sophisticated method \cite{Johansson2017} implemented multiple shape models: points, polygons, L-shape and lines for various objects. When tracking a moving object, its shape varies with the changes of pose and sensor viewpoint. \cite{Stefan2018} implemented a tracking method that simultaneous estimate the states of both poses and shapes represented by 2D polylines. 

\subsection{Object Intention Prediction}
The previous introduced modules provide the past and current information of detected targets. While in autonomous driving systems, decision making and path planning algorithms require future motion of the tracked targets. Prior works based on certain kinematic models that are assumed to perfectly fit the detected objects, are not applicable for long term prediction. To address this shortcoming, maneuver or behavior recognition is proposed based on machine learning methods. Common maneuvers for a vehicle are cut-in, lane-change, brake, over-taking, etc. \cite{spmBehavior2016} modeled the behaviors of car-following, lane-change by GMM (Gaussian mixture model) or HMM (Hidden Markov Model). Based on the maneuver classification realized by HMM, \cite{Deo2018} predicted the vehicles' motions by VGMMs (variational Gaussian mixture models) under the constraints of vehicle interaction models. With the success of RNN in modeling temporal sequential data, LSTM based methods are becoming popular. \cite{Derek2017} used LSTM to classify the drivers' intentions at intersections, and the results shown that LSTM outperforms other traditional machine learning methods. \cite{deoIV2018} proposed a encoder-decoder LSTM model to recognize maneuver and predict trajectory. Beyond recognizing the maneuver of a single object, \textit{social LSTM} \cite{socialLSTM2016} was proposed to capture the interactions of all the objects. This is achieved by \textit{social pooling}, which down-samples a target's neighboring objects' LSTM states into a \textit{social tensor}. \cite{deoCVPR2018} and \cite{mess2019} applied and improved original social pooling part for the purpose of vehicle trajectory prediction.  


\subsection{Emerging Deep Learning Methods}
After the huge success in computer vision and speech recognition, waves of deep learning (DL) arrived in LiDAR data processing as well. Deep learning \cite{dlMIT2016} is a subset of machine learning algorithms that mainly uses multi-layer neural network. In contrast to the traditional machine learning methods such as SVM, DL technologies are able to automatically extract features from the raw input. Convolutional neural networks (CNN) and recurrent neural networks (RNN), such as LSTM (long short-term memory), are the most frequently used tools.

The basic components of perception system, \textit{ground segmentation, object detection, tracking and recognition} can all be realized by deep neural networks (DNN). For instance, \cite{martin2018} segmented ground points by applying CNN to LiDAR points represented by multi-channel range images. In contrast to the object detection based on clustering that arbitrary targets can be detected, DNN based solutions achieve object detection by recognition, owning to the paradigm of supervised learning. As in \cite{pixor2018}, vehicles can be detected by CNN based neural network in a BEV (bird's eye view) representation of LiDAR points. A more complex neural network was proposed in \cite{mv3d2017}, CNN is utilized in both range image and BEV of LiDAR data, and then fused with camera detections. However, due to the physical limits of LiDAR, only vehicles can be effectively detected by LiDAR, the best achieved results of pedestrian detection in KITTI benchmark is only $52.40\%$ in average precision (by method DENFIDet when writing this paper) \footnote{\url{http://www.cvlibs.net/datasets/kitti/eval_object.php?obj_benchmark=bev}}. \cite{Edouard2019} integrated the evidential theory into DL architecture for LiDAR based road segmentation and mapping. Object tracking has been realized by DL as well. In contrast to the \textit{tracking by filtering} framework described in traditional tracking algorithm, \cite{dnnTracking2018} proposed a deep structure model under \textit{tracking by detection} framework. A detection net first process a sequence of LiDAR data and images to generate detection proposals. Then, tracks are estimated through finding best associations of detections, which is achieved by a marching net and scoring net.

\begin{figure}[t]
\centering
\includegraphics[width = 0.7\textwidth]{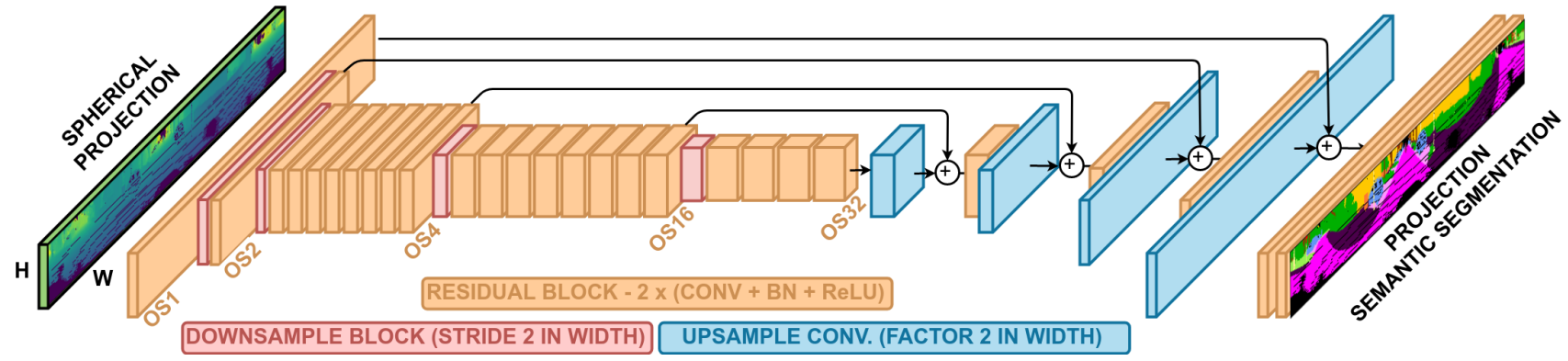}
\caption{The neural network structure proposed in RangeNet \cite{RangeNet}.}
\label{fig::rangenet}
\end{figure}

Apart from improving the tradition perception components, point-wise semantic segmentation, which was hard to be realized before, is now achievable by deep learning. A well-known method PointNet was proposed in \cite{PointNet2018} to semantically segment 3D point clouds for general usages. However, due to sparsity of LiDAR data w.r.t distance, the method doesn't work well for autonomous driving scenarios. SqueezeSeg \cite{SqueezeSeg} achieved real-time segmentation by applying CNN in range view of LiDAR points. Due to a lack of massive annotated datasets, the performances of these two methods are not ready for deployment in real usage. While this situation has been changed by SemanticKITTI \footnote{\url{http://www.semantic-kitti.org/}} \cite{skitti2019} -- the latest and the biggest point-wise annotated dataset based on KITTI. Based on this dataset, RangeNet \cite{RangeNet} demonstrated fascinating performance and speed from a not complicated DNN structure. Fig. \ref{fig::rangenet} shows the structure and a sample result of RangeNet. With more and more annotated datasets, we have sufficient reason to expect LiDAR based semantic segmentation will have better performance.

\section{Conclusion and Future Directions}\label{sec::conclusion}
In this article, a review of LiDAR technologies is presented in the beginning. How a LiDAR "sees" the world and what constitutes a LiDAR are introduced. The main development directions of LiDAR technologies are analyzed as well. In summary, current automotive LiDARs face the constraints, or challenges as: 1) cost, 2) meet automotive reliability and safety standards (e.g. ISO26262, IEC61508), 3) long measuring distance (e.g. $>$200m for highway applications), 4) adverse weather i.e. rain, fog, snow etc, 5) image-level resolution, 6) smaller size facilitating integration. At present, all the possible solutions varying from laser sources (905nm V.S. 1550nm), scanning methods (Spinning/MEMS/OPA/Flash) or ranging principles (ToF or FMCW), are exploited to overcome several or all of these difficulties. It is really hard to predict which automotive LiDAR solution(s) will dominate the future, whereas one thing is sure: automotive LiDARs are walking out of experimental platforms, entering more and more mass-produced cars. 

Then, a compact tutorial of the LiDAR based perception systems for autonomous driving is presented. Three levels of information providing by perception systems, along with typical processing pipeline are introduced. Generally, comparing to camera or radar, LiDAR is the most precise sensor in measuring range. Therefore, the physical information (objects' positions, headings, shapes, etc.) evaluated by LiDAR based algorithms is highly reliable. However, semantic description is the shortcoming of LiDAR. This is caused by the LiDAR's poor resolution and its essence as a distance measuring sensor, not contextual sensor. Fusion with cameras remedies LiDAR's weakness in recognition. The intention prediction level is independent of specific sensor, while it will be strengthened by the precise physical information bringing by LiDAR. Applying deep learning in LiDARs' 3D data will be one of the most important directions in the future. Lacking huge amount of annotated 3D point cloud dataset was the bottleneck of successfully applying deep learning methods. However, things are changing. The aforementioned SemanticKITTI initiated a good start, and the results achieved by RangeNet++ are quite impressive. From the authors' point of view, algorithms for extracting more accurate physical information and squeezing the LiDARs' potentials in semantic estimation are the future directions. And of cause, with the fast progress of new LiDARs, new algorithms will emerge adapted to specific LiDARs.


\vspace{-0.2cm}
\bibliographystyle{IEEEtran}
\bibliography{IEEEabrv,refs}

\end{document}